\documentclass[runningheads]{llncs}
\usepackage[T1]{fontenc}
\usepackage[utf8]{inputenc}
\usepackage{graphicx,verbatim}
\usepackage{booktabs}
\usepackage{multirow}
\usepackage{amsmath}
\usepackage{amssymb}
\usepackage{booktabs}
\usepackage{multirow}

\usepackage[colorlinks]{hyperref}
\usepackage{color}
\usepackage{xcolor}
\hypersetup{colorlinks,
allcolors=black}

\begin{document}
\title{Genetically Aligned Patient Representations Improve Hematological Diagnosis}
\titlerunning{GenBloom}
%
\author{Muhammed Furkan Dasdelen\inst{1,2}\thanks{These authors contributed equally.} \and
Fatih Ozlugedik\inst{1}\textsuperscript{*} \and
Ilaria Looser\inst{1} \and
Rao Muhammad Umer\inst{1} \and
Christian Pohlkamp\inst{3} \and
Carsten Marr\inst{1,4,5,6,7}\thanks{Correspondence: carsten.marr@helmholtz-munich.de}}
\authorrunning{Dasdelen et al.}
%
\institute{Institute of AI for Health, Helmholtz Munich, Germany \and International School of Medicine, Istanbul Medipol University, T\"urkiye \and
Munich Leukemia Laboratory, Germany \and
Department of Medicine III, Ludwig-Maximilian-University Hospital, Germany \and
Department of Physics, University of Munich, Germany \and
Munich Center for Machine Learning (MCML), Germany \and
DKTK, German Cancer Consortium, Germany}

\maketitle              
\begin{abstract}
Multimodal alignment of histopathology encoders with transcriptomic and genomic data has been shown to significantly improve performance in downstream diagnostic tasks. Hematological cytology is unique in that visual single-cell evaluation is often paired with cytogenetics and molecular genetics for blood cancer diagnosis. In this study, we present a framework to align single white blood cell images with chromosomal aberrations (karyotype) and somatic mutations from targeted gene panels. Our training strategy follows a two-stage approach: (i) self-supervised, vision-only pretraining of a transformer aggregator using an iBOT head on a cohort of over 1500 patients, and (ii) genetic alignment via supervised contrastive loss on acute myeloid leukemia patients. Our genetically aligned patient encoder improves hematological diagnostic tasks, outperforming slide-level histopathology foundation models. Additionally, the model provides off-the-shelf retrieval capabilities for diseases and genetic alterations. Incorporating genetic data into patient encoders increases the quality of patient representations, providing a framework that aligns with clinical diagnostic workflows and paves the way for future multimodal hematology-specific AI. The code and model weights are available at \url{https://github.com/marrlab/GenBloom}.

\keywords{cytology  \and karyotype \and multimodal alignment \and genetic.}

\end{abstract}

\section{Introduction}

Hematological diagnosis fundamentally relies on integrating microscopic morphology with cytogenetic and molecular profiling for precise tumor classification, risk stratification, and treatment selection \cite{khoury20225th,jaffe2001pathology,vardiman2002world}. While evaluating peripheral blood and bone marrow smears provides a rapid initial assessment, genetic profiling is essential to formally classify the disease. This tight clinical coupling of modalities motivates the need for joint computational modeling.

In computational pathology, deep learning foundation models trained on whole-slide images (WSIs) have shown strong performance on downstream tasks like tumor subtyping and survival prediction \cite{xu2024whole,shaikovski2024prism,ding2025multimodal}. Recent multimodal frameworks further demonstrate that aligning histopathology representations with molecular data improves predictive performance, generalization, and the biological relevance of visual embeddings \cite{vaidya2025molecular,wang2026transcriptomic}. These findings confirm that morphology and molecular features share an exploitable latent space.

Blood smears provide information at single-cell resolution—the scale where morphological variation most directly reflects underlying genetic alterations. Existing hematological models predominantly analyze morphology alone \cite{hehr2023explainable,dasdelen2026ai,sidhom2021deep,eckardt2022deep}, failing to capture molecular heterogeneity. Furthermore, the scarcity of large-scale paired datasets restricts the development of models.

To bridge this gap, we present GenBloom, the first genetically aligned, slide-level blood model tailored to hematology. GenBloom integrates single white blood cell images with cytogenetic abnormalities and somatic mutations to learn comprehensive patient embeddings. Our two-stage training paradigm involves: (i) large-scale self-supervised pretraining on cellular morphology to extract robust visual features, and (ii) supervised multimodal alignment to anchor patient embeddings within the genetic space. This design successfully captures clinically meaningful relationships between cellular morphology and molecular disease drivers.

\section{Methods}
\subsection{Pretraining dataset}
\begin{figure}[t]
\includegraphics[width=\textwidth]{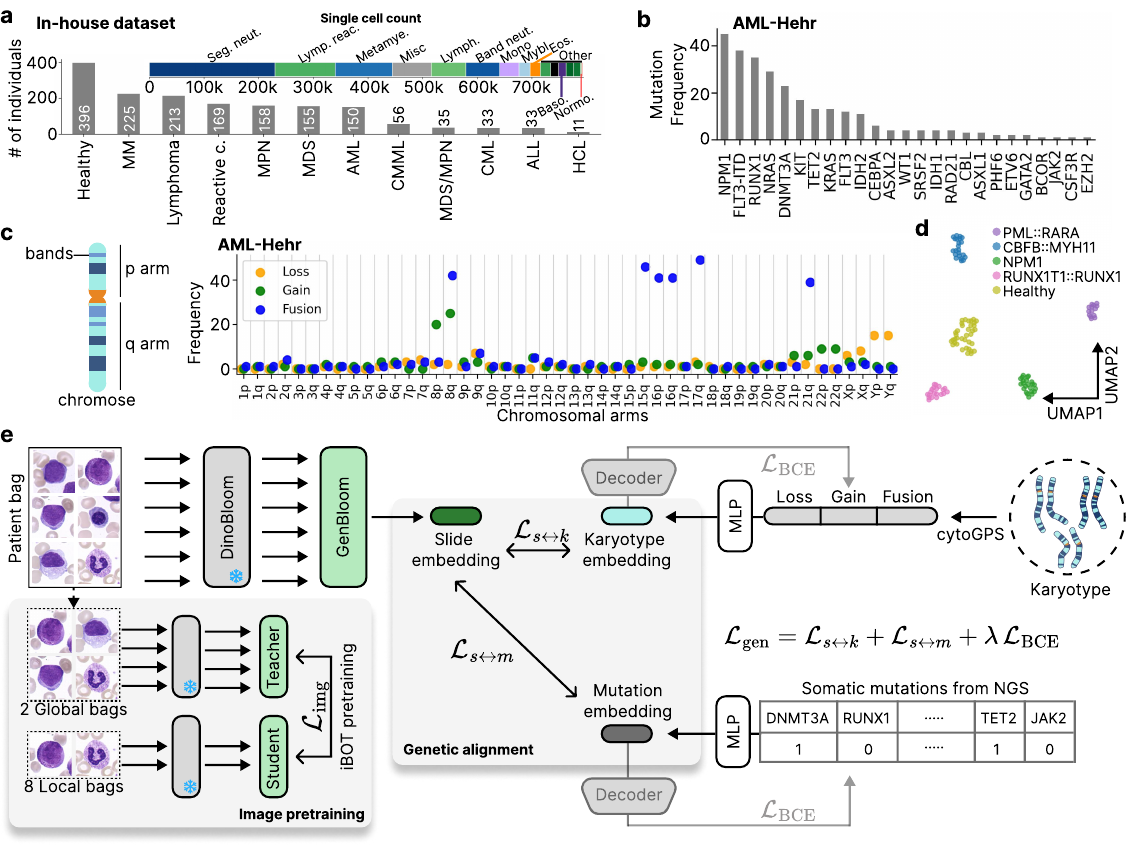}
\caption{\textbf{GenBloom pretraining and genetic alignment.} \textbf{(a)} Image pretraining cohort contains $>$1,500 patients and $>$700k single-cell images spanning major hematologic entities and cell lineages. \textbf{(b)} AML-Hehr mutation frequencies for genes used in the alignment. \textbf{(c)} Distribution of loss, gain and fusion events across chromosomal arms in AML-Hehr. \textbf{(d)} UMAP of GenBloom-G embeddings on AML-Hehr training patients colored by recurrent cytogenetic/molecular subtypes. \textbf{(e)} DINOv2-adapted image pretraining and supervised contrastive genetic alingment for GenBloom.}\label{fig1}
\end{figure}

    For image pretraining, we used an in-house peripheral blood smear dataset (collected at Munich Leukemia Laboratory), which contains single-cell images from 1,634 patients spanning a range of hematological diseases (acute leukemia, myelodysplastic syndrome (MDS), myeloproliferative neoplasm (MPN), overlap syndromes, lymphoma, multiple myeloma, and reactive changes) as well as healthy controls. It includes 794,527 single-cell images for pretraining (Fig.~\ref{fig1}a). For genetic alignment, we used the AML-Hehr dataset \cite{hehr2023explainable}, which comprises peripheral blood smear images from 189 acute myeloid leukemia (AML) patients and a healthy cohort, paired with molecular and cytogenetic profiles. The dataset includes 37 unique somatic mutations and 52 distinct karyotypes (Fig.~\ref{fig1}b, c). We held out a test set (n=43) and contrastively aligned the remaining 146 patients’ images with their genetic information.

\subsection{Evaluation tasks and dataset}
    For downstream evaluation, we considered three publicly available patient-level classification tasks. First, we performed AML genetic subtyping on the held-out AML-Hehr test set, focusing on \textit{PML::RARA} fusion (train/test $n=18/6$), \textit{CBFB::MYH11} fusion (train/test $n=28/9$), \textit{NPM1} mutation (train/test $n=28/8$), and \textit{RUNX1::RUNX1T1} fusion (train/test $n=25/7$), and controls (train/ test $n=47/13$). The other two datasets are out-of-domain and were used to test generalizability: the APL-AML dataset includes acute promyelocytic leukemia (APL, train/test $n=22/12$) and other acute myeloid leukemias (AML, train/test $n=60/12$), and AMH was curated from the cAItomorph test set \cite{dasdelen2026ai}, including AML (train/test $n=79/20$) and healthy individuals (train/test $n=29/8$). In all experiments, we performed 5-fold cross-validation while keeping the test set fixed.
    
    We also performed retrieval analysis. In retrieval experiments, we embedded each patient's slide and genomic profile (karyotype or mutation) into a shared representation space and performed cosine similarity to rank candidates. We evaluated genomics$\rightarrow$slide by retrieving the correct slide given a genomic query, slide$\rightarrow$genomics by retrieving the correct genomic profile given a slide query, and slide$\rightarrow$slide by retrieving slides from the same disease given a slide query.
    
\subsection{Data processing}
    Single cell images were reshaped to $224\times224$, normalized with Imagenet statistics. We used DinoBloom-B \cite{koch2024dinobloom} hematology image encoder to create single cell representations. DinoBloom was frozen throughout all experiments. 

    Structural abnormalities were characterized through cytogenetics and fluorescence in situ hybridisation, with karyotyping by chromosome banding analysis documented according to the International System for Human Cytogenetic Nomenclature (ISCN) standards \cite{schoch2002comparison,mcgowan2020iscn}. Karyotype data were processed using CytoGPS \cite{abrams2019cytogps}, which converts ISCN text strings into patient-level binary indicators of chromosomal loss, gain, and fusion. Encoding three indicators per cytoband (368 bands) yielded a 1,104-dimensional input for the cytogenetics branch of the model (Fig.~\ref{fig1}c,e).
    
    At diagnosis, patients also underwent targeted molecular genetics assessment following the protocol previously described \cite{fuhrmann2022aml}. Pathogenic variants were aggregated at the gene level, collapsing all alterations of a given gene into a single binary indicator. We retained features with recorded measurements for at least 30 patients and with both positive and negative labels present, yielding 25 binary gene-level mutation features for the molecular genetics branch (Fig.~\ref{fig1}b,e).

\subsection{Image pretraining}
    GenBloom is a patient-level transformer aggregator \cite{dosovitskiy2020image} (vision transformer, ViT) with $L{=}6$ layers, $H{=}12$ heads, embedding dimension $D{=}768$ (MLP hidden dim $3072$). This small ViT has been shown to effectively encode slide-level information in histopathology images \cite{ding2025multimodal}. GenBloom operates on an unordered set of single-cell embeddings; therefore, we removed patchification (replaced by an MLP) and positional encodings of ViT to enforce permutation invariance. For each patient with up to $500$ cells, we extracted per-cell embeddings using a frozen DinoBloom-B \cite{koch2024dinobloom} encoder and aggregated them with GenBloom, using the \texttt{[CLS]} token as the patient representation.
    
    We adapted DINOv2/iBOT pretraining \cite{oquab2023dinov2,zhou2021ibot} to patient bags (or slides) via multi-crop subsampling. Our training pipeline operates in embedding space rather than on raw images (Fig.~\ref{fig1}d). For each patient, we sampled $K_g{=}2$ global bags and $K_\ell{=}8$ local bags by randomly selecting $70\%$ ($\approx 350$ cells) and $20\%$ ($\approx 100$ cells) of cell embeddings, respectively. We trained a student ($s$)--teacher ($t$) model, where the teacher is an exponential moving average (EMA) of the student. The objective combines (i) DINO-style \texttt{[CLS]} alignment across views and (ii) iBOT masked embedding prediction on randomly masked cell embeddings:
    \[
    \mathcal{L}_{\text{img}}=\mathcal{L}_{\text{DINO}}+\lambda\,\mathcal{L}_{\text{iBOT}},
    \]
    \[
    \mathcal{L}_{\text{DINO}}
    =\frac{1}{|\mathcal{P}|}\sum_{(g,k)\in\mathcal{P}}
    \mathrm{CE}\!\left(p_t^{(g)},\,p_s^{(k)}\right),
    \qquad
    \mathcal{L}_{\text{iBOT}}
    =\frac{1}{|\mathcal{M}|}\sum_{i\in\mathcal{M}}
    \mathrm{CE}\!\left(q_i^{t},\,q_i^{s}\right),
    \]
    where $\mathcal{P}$ denotes all (teacher global, student view) pairs, $\mathcal{M}$ the set of masked tokens, and $\mathrm{CE}(a,b)=-\sum_{c} a_c \log b_c$. This pretraining encourages GenBloom to learn robust patient representations and to model cell-composition statistics under realistic subsampling.
    
    We trained GenBloom on a single NVIDIA H100 80GB GPU for 100 epochs with 64 batch size ($\approx 7.26\,\mathrm{kg}\ \mathrm{CO}_2\text{eq}$ emitted for all experiments).

\begin{figure}[t]
\includegraphics[width=\textwidth]{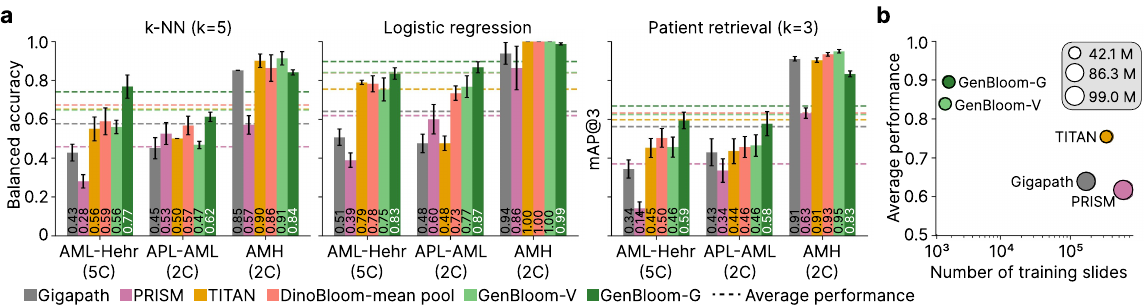}
\caption{\textbf{GenBloom outperforms histopathology slide encoders on hematology tasks.} \textbf{(a)} Performance on AML-Hehr (5-class), APL-AML (2-class), and AMH (2-class) using $k$-NN probing ($k{=}5$), logistic regression, and patient retrieval (mAP@$3$). \textbf{(b)} Average performance versus pretraining scale (number of training slides; marker size indicates parameter count)} \label{fig2}
\end{figure}

\subsection{Genetic alignment}
 After image pretraining, we performed supervised genetic alignment in embedding space to couple morphology with cytogenetics and molecular genetics (Fig.~\ref{fig1}d). For each patient $p$, GenBloom produced a slide-level embedding from the \texttt{[CLS]} token, $s_p\in\mathbb{R}^{768}$. We encoded cytogenetics (karyotype) and molecular genetics (mutations) as binary vectors $y^{k}_p\in\{0,1\}^{d_k}$ and $y^{m}_p\in\{0,1\}^{d_m}$, respectively (${d_k}=1,104$, ${d_m}=25$).

Modality-specific MLP projection heads mapped all modalities into a shared 128-dimensional $\ell_2$-normalized space:
\[
z^s_p = \frac{\phi_s(s_p)}{\|\phi_s(s_p)\|},\qquad
z^k_p = \frac{\phi_k(y^k_p)}{\|\phi_k(y^k_p)\|},\qquad
z^m_p = \frac{\phi_m(y^m_p)}{\|\phi_m(y^m_p)\|}, \qquad z^\cdot_p\in\mathbb{R}^{128}.
\]

We aligned modalities using a cross-modal supervised contrastive objective~\cite{khosla2020supervised}. For anchor modality $a$ and target modality $b$, let $\mathcal{P}(p)=\{j \neq p : c_j = c_p\}$ denote the set of samples in the batch (with batch size $B$) sharing the same class label ($c_p$) as patient $p$, excluding the anchor itself. The unidirectional loss is:
\[
\mathcal{L}_{a\to b}
= -\frac{1}{B}\sum_{p=1}^{B}
\frac{1}{|\mathcal{P}(p)|}
\sum_{j\in\mathcal{P}(p)}
\log
\frac{\exp\!\left(z^a_p{}^\top z^b_j\,/\,\tau\right)}
{\sum_{q=1}^{B}\exp\!\left(z^a_p{}^\top z^b_q\,/\,\tau\right)},
\]
where $\tau$ is a temperature parameter and the dot product equals cosine similarity due to $\ell_2$-normalization. Each alignment is trained symmetrically: $\mathcal{L}_{a\leftrightarrow b} = \frac{1}{2}(\mathcal{L}_{a\to b} + \mathcal{L}_{b\to a})$.

To preserve modality-specific biological information and reduce representational collapse, we attached lightweight decoders to the bottleneck representations. These decoders reconstructed the original binary genetic vectors, where each entry indicates the presence or absence of a cytogenetic or molecular alteration. Reconstruction was supervised with a binary cross-entropy loss $\mathcal{L}_{\text{BCE}}$, which independently penalizes incorrect predictions for each genetic feature. The total objective was:

\[
\mathcal{L}_{\text{gen}}
= \mathcal{L}_{s\leftrightarrow k}
+ \mathcal{L}_{s\leftrightarrow m}
+ \lambda_r\,\mathcal{L}_{\text{BCE}},
\]
where $\lambda_r$ controls the reconstruction regularizer strength. This training aligned slide, karyotype, and mutation representations in a shared space, enabling retrieval and downstream prediction from any modality.

\subsection{Baseline models}
    We compared GenBloom against 3 histopathology slide encoders—GigaPath \cite{xu2024whole}, PRISM \cite{shaikovski2024prism}, and TITAN \cite{ding2025multimodal}—using linear probing. We also added another baseline by simply averaging DinoBloom embeddings (mean pooling). GigaPath (86.3M param.) was trained on 171,189 H\&E-stained whole-slide images (WSIs). TITAN (42.1M param.) was trained on 335,645 H\&E- and IHC-stained WSIs with report alignment. PRISM (99M param.) was trained on 587,196 WSIs with report alignment. For linear probing, we used logistic regression (\textit{lbfgs} solver, regularization coefficient $C{=}1$) and $k$-NN implemented in \textit{scikit-learn}.

\begin{table}[t]
\caption{In-domain cross-modal retrieval performance of GenBloom-G on the AML-Hehr test set. We report top-1, top-5 accuracy and mean reciprocal rank (MRR) for slide$\leftrightarrow$karyotype (S$\leftrightarrow$K) and slide$\leftrightarrow$mutation (S$\leftrightarrow$M) retrieval. $^{*}p < 0.001$.}\label{tab:retrieval}
\centering
\scriptsize
\setlength{\tabcolsep}{4pt}
\begin{tabular}{@{}l c cc cc@{}}
\toprule
& & \multicolumn{2}{c}{\textbf{S$\leftrightarrow$K}} & \multicolumn{2}{c}{\textbf{S$\leftrightarrow$M}} \\
\cmidrule(lr){3-4} \cmidrule(lr){5-6}
\textbf{Metric} & \textbf{Direction} & GenBloom-G & Random & GenBloom-G & Random \\
\midrule
\multirow{2}{*}{Top-1 Acc}
 & $\to$       & 0.09{\tiny$\pm$0.04}$^{*}$ & 0.05{\tiny$\pm$0.04} & 0.12{\tiny$\pm$0.04}$^{*}$ & 0.06{\tiny$\pm$0.05} \\
 & $\leftarrow$ & 0.14{\tiny$\pm$0.03}$^{*}$ & 0.05{\tiny$\pm$0.04} & 0.12{\tiny$\pm$0.02}$^{*}$ & 0.06{\tiny$\pm$0.05} \\
\midrule
\multirow{2}{*}{Top-5 Acc}
 & $\to$       & 0.62{\tiny$\pm$0.06}$^{*}$ & 0.23{\tiny$\pm$0.09} & 0.46{\tiny$\pm$0.06}$^{*}$ & 0.29{\tiny$\pm$0.08} \\
 & $\leftarrow$ & 0.55{\tiny$\pm$0.04}$^{*}$ & 0.23{\tiny$\pm$0.09} & 0.66{\tiny$\pm$0.08}$^{*}$ & 0.30{\tiny$\pm$0.11} \\
\midrule
\multirow{2}{*}{MRR}
 & $\to$       & 0.41{\tiny$\pm$0.04}$^{*}$ & 0.20{\tiny$\pm$0.05} & 0.30{\tiny$\pm$0.03}$^{*}$ & 0.21{\tiny$\pm$0.05} \\
 & $\leftarrow$ & 0.33{\tiny$\pm$0.03}$^{*}$ & 0.17{\tiny$\pm$0.04} & 0.33{\tiny$\pm$0.02}$^{*}$ & 0.20{\tiny$\pm$0.05} \\
\bottomrule
\end{tabular}
\end{table}

\subsection{Statistical analysis and metrics}
    Balanced accuracy (bAcc) was used for classification tasks, mean average precision (mAP) at $k=3$ for slide-to-slide retrieval, and the F1 score for cross-modal retrieval. We performed 1{,}000 bootstrap iterations, randomly resampling the test set for retrieval tasks, and used the Wilcoxon signed-rank test for statistical comparisons, with a Bonferroni correction. Corrected $p$-values below 0.05 were considered statistically significant.

\section{Results}

\subsection{GenBloom improves hematological classification}
We first evaluated GenBloom on downstream classification tasks using $k$-NN ($k{=}5$) and logistic regression (Fig.~\ref{fig2}a). We denote the model after image pretraining as GenBloom-V and the model after genetic alignment as GenBloom-G. 


On the AML-Hehr genetic subtyping task (5 classes), GenBloom-G achieved the highest balanced accuracy, outperforming the second best model TITAN by 38\% in $k$-NN and 5\% in logistic regression. On APL-AML dataset (2 classes), GenBloom-G outperformed the second best histopathology slide encoder PRISM by 15\% in $k$-NN and by 45\% in logistic regression. On AML vs. healthy (AMH), GenBloom-V was marginally better than TITAN in $k$-NN, with no difference in logistic regression. Overall, GenBloom-G ranked significantly higher than the other models across datasets for both $k$-NN and logistic regression (Friedman test, $p < 0.001$ for both).


We also performed a retrieval analysis to assess how well the model retrieves clinically relevant patients with the same diagnosis for a given query patient, using $k=3$. GenBloom-G achieved the highest mAP@3, followed by GenBloom-V, on the AML-Hehr test set and the APL-AML dataset, outperforming the next-best histopathology slide encoder (TITAN) by $\approx31\%$. On the AMH dataset, GenBloom-V achieved the best performance, followed by TITAN. 


Notably, GenBloom achieved these results with substantially fewer parameters and less training data than the other slide-level models (Fig.~\ref{fig2}b), highlighting the importance of domain-specific pretraining. Surprisingly, mean pooling of DinoBloom embeddings achieved the second- or third-best performance in linear probing and retrieval tasks on average, showing the advantage of using a hematology-specific image encoder. 

\begin{table}[t]
\caption{Out-of-domain per-gene retrieval F1 score on the cAItomorph cohort for slide-to-mutation (S$\to$M) and mutation-to-slide (M$\to$S) directions. $N$: number of positive patients. Fold difference: ratio of GenBloom F1 to random F1. $^{*}p < 0.001$.}\label{tab:pergene_ood}
\centering
\scriptsize
\setlength{\tabcolsep}{2.5pt}
\begin{tabular}{@{}l r ccc ccc@{}}
\toprule
& & \multicolumn{3}{c}{\textbf{S$\to$M}} & \multicolumn{3}{c}{\textbf{M$\to$S}} \\
\cmidrule(lr){3-5} \cmidrule(lr){6-8}
\textbf{Gene} & $N$ & GenBloom-G & Random & Fold diff. & GenBloom-G & Random & Fold diff. \\
\midrule
\textit{ASXL1}    &  81 & 0.23{\tiny$\pm$0.08}$^{*}$ & 0.16{\tiny$\pm$0.04} & 1.47 & 0.41{\tiny$\pm$0.14}$^{*}$ & 0.16{\tiny$\pm$0.04} & 2.56 \\
\textit{DNMT3A}   &  96 & 0.29{\tiny$\pm$0.03}$^{*}$ & 0.19{\tiny$\pm$0.04} & 1.53 & 0.23{\tiny$\pm$0.11}$^{*}$ & 0.19{\tiny$\pm$0.04} & 1.23 \\
\textit{FLT3-ITD} &  11 & 0.08{\tiny$\pm$0.07}$^{*}$ & 0.02{\tiny$\pm$0.05} & 4.10 & 0.01{\tiny$\pm$0.03} & 0.02{\tiny$\pm$0.05}$^{*}$ & 0.67 \\
\textit{IDH2}     &  22 & 0.14{\tiny$\pm$0.03}$^{*}$ & 0.04{\tiny$\pm$0.05} & 3.16 & 0.19{\tiny$\pm$0.08}$^{*}$ & 0.04{\tiny$\pm$0.05} & 4.48 \\
\textit{JAK2}     & 213 & 0.57{\tiny$\pm$0.04}$^{*}$ & 0.42{\tiny$\pm$0.03} & 1.36 & 0.55{\tiny$\pm$0.23}$^{*}$ & 0.42{\tiny$\pm$0.03} & 1.33 \\
\textit{NPM1}     &  12 & 0.08{\tiny$\pm$0.04}$^{*}$ & 0.02{\tiny$\pm$0.05} & 3.30 & 0.10{\tiny$\pm$0.08}$^{*}$ & 0.02{\tiny$\pm$0.05} & 4.10 \\
\textit{NRAS}     &  17 & 0.10{\tiny$\pm$0.05}$^{*}$ & 0.03{\tiny$\pm$0.05} & 2.95 & 0.35{\tiny$\pm$0.19}$^{*}$ & 0.03{\tiny$\pm$0.05} & 10.63 \\
\bottomrule
\end{tabular}
\end{table}

\subsection{Genetic alignment enables cross-modal retrieval}

To assess whether GenBloom-G learns a shared embedding space across modalities, we evaluated cross-modal retrieval between slide embeddings (S), karyotype embeddings (K), and mutation embeddings (M) on the held-out AML-Hehr test set (in-domain) and the cAItomorph cohort (out-of-domain). We report top-1, top-5 accuracy and mean reciprocal rank (MRR), comparing GenBloom-G against a random baseline across 1{,}000 bootstrap iterations (Table~\ref{tab:retrieval}).

GenBloom-G significantly outperformed the random baseline across all retrieval directions and metrics ($p < 0.001$). For slide-to-karyotype retrieval (S$\to$K), the model achieved a top-5 accuracy of $0.62 \pm 0.06$, compared to $0.23 \pm 0.09$ for the random baseline, representing a $2.7\times$ improvement. Karyotype-to-slide retrieval (K$\to$S) showed a similar pattern, with MRR increasing from $0.17$ to $0.33$. Slide-to-mutation (S$\to$M) and mutation-to-slide (M$\to$S) retrieval followed a comparable trend, with top-5 accuracy reaching $0.46$ and $0.66$, respectively. 


In the out-of-domain setting (Table~\ref{tab:pergene_ood}), we evaluated genes that are important for diagnosis and prognosis (\textit{NPM1}, \textit{FLT3-ITD}, \textit{ASXL1}, \textit{NRAS}) as well as treatment decisions (\textit{IDH2}, \textit{JAK2}). GenBloom successfully retrieved relevant embeddings in both directions (mutation-to-slide and slide-to-mutation), significantly outperforming the random baseline in 13/14 cases. Notably, 186 of the 213 patients with JAK2 mutations were diagnosed with myeloproliferative neoplasms.

\subsection{Ablation studies}

\begin{table}[t]
\caption{Ablation studies on the AML-Hehr test set. The best results per ablation are in \textbf{bold}.}
\label{tab:ablation}
\centering
\scriptsize
\setlength{\tabcolsep}{3pt}
\begin{tabular}{@{}ll ccc@{}}
\toprule
\textbf{Ablation} & \textbf{Setting} & \textbf{LogReg bAcc} & \textbf{S$\to$K MRR} & \textbf{K$\to$S MRR} \\
\midrule
\multirow{3}{*}{\scriptsize Aggregator}
 & Transformer (GenBloom-V init.) & \textbf{0.83{\tiny$\pm$0.03}} & \textbf{0.38{\tiny$\pm$0.10}} & \textbf{0.22{\tiny$\pm$0.03}} \\
 & Transformer (Random init.)     & 0.82{\tiny$\pm$0.03}          & 0.36{\tiny$\pm$0.09}          & 0.21{\tiny$\pm$0.05} \\
 & Mean pooling                   & 0.78{\tiny$\pm$0.04}          & 0.28{\tiny$\pm$0.03}          & 0.19{\tiny$\pm$0.02} \\
\midrule
\multirow{2}{*}{\scriptsize\shortstack{Karyotype\\resolution}}
 & Band-level             & \textbf{0.83{\tiny$\pm$0.03}} & \textbf{0.38{\tiny$\pm$0.10}} & 0.22{\tiny$\pm$0.03} \\
 & Arm-level              & 0.81{\tiny$\pm$0.04}          & 0.35{\tiny$\pm$0.08}          & \textbf{0.23{\tiny$\pm$0.04}} \\
\midrule
\multirow{3}{*}{\scriptsize $\lambda_r$}
 & $\lambda_r = 1.0$        & \textbf{0.83{\tiny$\pm$0.03}} & \textbf{0.38{\tiny$\pm$0.10}} & \textbf{0.22{\tiny$\pm$0.03}} \\
 & $\lambda_r = 0.1$        & 0.80{\tiny$\pm$0.03}          & 0.25{\tiny$\pm$0.07}          & 0.16{\tiny$\pm$0.03} \\
 & $\lambda_r = 0.0$        & 0.81{\tiny$\pm$0.06}          & 0.28{\tiny$\pm$0.10}          & 0.19{\tiny$\pm$0.04} \\
\bottomrule
\end{tabular}
\end{table}

We ablated three design choices in the genetic alignment stage---the vision aggregator, karyotype encoding resolution, and reconstruction loss weight for genetic input $\lambda_r$---on the AML-Hehr test set (Table~\ref{tab:ablation}). Finetuning the pretrained transformer (GenBloom-V initialized) achieved the best overall performance (bAcc $0.83$, S$\to$K MRR $0.38$, K$\to$S MRR $0.22$) compared to a randomly initialized transformer and mean pooling. Band-level karyotype encoding consistently outperformed the coarser arm-level representation (bAcc $0.83$ vs.\ $0.81$; S$\to$K MRR $0.38$ vs.\ $0.35$), indicating that finer cytogenetic resolution provides a more informative alignment signal. Finally, the reconstruction objective improved S$\to$K MRR by 36\%, K$\to$S MRR by 16\% and logistic regression bAcc by 3\%.

\section{Conclusion}
We developed GenBloom, a hematology-specific slide-level encoder that learns genetically aligned patient representations. It outperforms large-scale histopathology foundation models on downstream hematology tasks, despite being trained on substantially less data, showing the benefits of domain-specific pretraining. Aligning morphology with karyotype and mutation profiles enable cross-modal retrieval between images and genetic information. This genetic alignment opens a path toward genetics-aware smear analysis, supporting faster triage, prioritization for confirmatory testing, and more informed treatment decisions in clinical workflows.



    

\begin{credits}
\subsubsection*{Author contributions} Conceptualization: CM, MFD, FO; Data curation: CP, IL, MFD; Methodology and software: FO, MFD, IL, RMU; Writing-original draft: MFD, FO, IL; Writing–editing: RMU, CM. 

\subsubsection{\ackname} CM received funding from the European Research Council under the European Union’s Horizon 2020 Research and Innovation Programme (grant agreements 866411, 101113551, and 101213822) and support from the Hightech Agenda Bayern.

\subsubsection{\discintname} The authors declare no competing interests.
\end{credits}

%
%
%
\bibliographystyle{splncs04}
\bibliography{references}




\end{document}